\documentclass[conference]{IEEEtran}
\IEEEoverridecommandlockouts
\usepackage{cite}
\usepackage{amsmath,amssymb,amsfonts}
\usepackage{algorithmic}
\usepackage{graphicx}
\usepackage{textcomp}
\usepackage{xcolor}
\usepackage{hyperref}
\usepackage{tabularx}
\usepackage{xcolor}
\usepackage{amsmath}
\usepackage{enumitem}
\usepackage[linesnumbered,ruled,vlined]{algorithm2e}
\SetKwInput{KwInput}{Input}                
\SetKwInput{KwOutput}{Output}              
\SetKwInput{KwBuffer}{Reservoir}              

\newcommand{\etal}{\textit{et al.~}}
\newcommand\ie{\textit{i.e.~}}
\newcommand\etc{\textit{etc.~}}

\def\BibTeX{{\rm B\kern-.05em{\sc i\kern-.025em b}\kern-.08em
    T\kern-.1667em\lower.7ex\hbox{E}\kern-.125emX}}
\begin{document}
\title{Attributes-aware Visual Emotion Representation Learning\\
}

\author{\IEEEauthorblockN{Anonymous Authors}}

\author{\IEEEauthorblockN{1\textsuperscript{st} Rahul Singh Maharjan}
\IEEEauthorblockA{\textit{Department of Computer Science} \\
\textit{University of Manchester}\\
Manchester, United Kingdom\\
rahulsingh.maharjan@manchester.ac.uk}
\and
\IEEEauthorblockN{2\textsuperscript{nd} Marta Romeo}
\IEEEauthorblockA{\textit{School of Mathematical \& Computer Sciences}\\
\textit{Heriot-Watt University}\\
Edinburgh, United Kingdom\\
M.Romeo@hw.ac.uk}
\and
\IEEEauthorblockN{3\textsuperscript{rd} Angelo Cangelosi}
\IEEEauthorblockA{\textit{Department of Computer Science} \\
\textit{University of Manchester}\\
Manchester, United Kingdom\\
angelo.cangelosi@manchester.ac.uk}
}

\maketitle
\begin{abstract}

Visual emotion analysis or recognition has gained considerable attention due to the growing interest in understanding how images can convey rich semantics and evoke emotions in human perception. However, visual emotion analysis poses distinctive challenges compared to traditional vision tasks, especially due to the intricate relationship between general visual features and the different affective states they evoke, known as the affective gap. Researchers have used deep representation learning methods to address this challenge of extracting generalized features from entire images. However, most existing methods overlook the importance of specific emotional attributes such as brightness, colorfulness, scene understanding, and facial expressions. Through this paper,  we introduce A4Net, a deep representation network to bridge the affective gap by leveraging four key attributes: brightness (Attribute 1), colorfulness (Attribute 2), scene context (Attribute 3), and facial expressions (Attribute 4). By fusing and jointly training all aspects of attribute recognition and visual emotion analysis, A4Net aims to provide a better insight into emotional content in images. Experimental results show the effectiveness of A4Net, showcasing competitive performance compared to state-of-the-art methods across diverse visual emotion datasets. Furthermore, visualizations of activation maps generated by A4Net offer insights into its ability to generalize across different visual emotion datasets.
\end{abstract}


\section{Introduction}\label{introduction}

Emotions represent diverse cognitive mechanisms that our minds utilize to enhance cognitive abilities \cite{minsky2007emotion}. Over recent years, there has been a noticeable trend toward expressing and sharing opinions and emotions online, employing various mediums, including text, images, and videos. Visual emotion analysis has accumulated significant attention, seeking to discern the emotional responses of individuals toward different visual stimuli. The comprehension of information within the expanding reservoir of data holds paramount importance for behavioral science \cite{fang2015relational}, which endeavors to forecast decision-making and facilitate applications including mental health assessment \cite{jiang2017learning, wieser2012reduced}, business recommendations \cite{mitchell1986effect}, and entertainment assistance \cite{chen2020adaptive}. Since emotions are inherent to human nature, artificial agents should strive to gain a deeper understanding of emotions to emulate human behavior more effectively.

In computer vision and affective computing research, visual emotion analysis is challenging due to the affective gap \cite{hanjalic2006extracting}, which indicates the absence of a proper connection between the features and the expected emotional state. Taking inspiration from psychological and art theory \cite{frijda1986emotions}, researchers manually created hand-crafted features consisting of color, text \etc \cite{machajdik2010affective}. Different from hand-crafted features, deep representation learning methods can extract emotional features automatically end-to-end. With the advancement of deep representation learning, the research focus on visual emotion analysis has shifted from traditional hand-crafted feature designing \cite{lu2012shape, zhao2014exploring} to deep representation learning \cite{peng2015mixed, rao2020learning, yang2017joint}. These deep representation learning methods usually focus on emotion classification without understanding the components of the images, such as color or scene. Regardless, most representation learning-based methods extract features from the entire image; however, they fail to evaluate the distinctive attributes of emotion evocation concerned in visual emotion analysis. 

Visual emotion analysis has received considerable attention in the field of psychology research as well. Frijda \cite{frijda1986emotions} suggested that certain objects and situations can elicit emotional responses. Brosch \etal \cite{brosch2010perception} conducted a review highlighting the significant role of emotional stimuli such as color, specific objects, facial expressions, or other attributes in perception and categorization. They emphasized that emotional categorization is a crucial mechanism through which humans organize their environment. Additionally, Brosch \etal \cite{brosch2010perception} noted that humans tend to infer semantic associations between scenes or objects depicted in images and specific emotions. Similarly, individuals often focus on facial expressions within images, which can evoke similar emotions \cite{brosch2010perception}. As such, Comprehending scenes or facial expressions during visual emotion analysis yields supplementary affective information. This additional information enables humans to extract enhanced features, improving visual emotion analysis capabilities.

Taking inspiration from previous works on the importance of brightness \cite{kurt2017modulation}, color 
 \cite{ritchie2013perceived}, understanding of scenes \cite{brosch2010perception}, and facial expression \cite{ekman1993facial} that evoke emotion, this paper aims to address the need for an attribute-aware visual emotion representation learning encompassing brightness (\textit{attribute 1}), colorfulness (\textit{attribute 2}), scene (\textit{attribute 3}), and facial expression (\textit{attribute 4}). We introduce A4Net, a deep representation attribute-aware visual emotion network designed to process input images and generate four distinct and rich feature vectors representing the emotion, colorfulness, brightness, scene, and facial expression depicted in the image. The output vector, particularly the emotion feature vector, holds potential for utilization in domain adaptation tasks \cite{wang2018deep} involving other visual emotion datasets.

To sum up, our contributions are as follows:
\begin{itemize}[noitemsep,topsep=0pt]
    \item We propose a novel attribute-aware visual emotion network, \ie, A4Net, that integrates four different image attributes to guide the network into learning rich emotion representation. 
    \item We performed thorough experiments on the EmoSet \cite{yang2023emoset}, EMOTIC \cite{kosti2019context}, SE30K8 \cite{wei2020learning}, and UnBiasEmo \cite{panda2018contemplating}. Our findings indicate that A4Net outperforms state-of-the-art methods across these datasets. Different from most previous work on visual emotion analysis, our results and visualization shed light on the importance of leveraging attributes such as color, brightness, scene, and facial expression to improve visual emotion analysis performance. 
\end{itemize}

\section{Related Work}
\subsection{Visual Emotion Analysis}

 For over two decades, researchers have been dedicated to analyzing emotions in visual images \cite{lang1997international, zhao2021affective}. Most existing approaches to visual emotion analysis can be categorized into either hand-crafted feature design or deep representation learning to reduce the \textit{affective gap} \cite{hanjalic2006extracting} (the gap between emotion and input visual). Earlier endeavors in visual emotion analysis predominantly focused on devising hand-crafted features. Machajdik and Hanbury \cite{machajdik2010affective} advocated using extracted low-level features like color and texture, combining them to predict the emotion. Yanulevskaya \etal \cite{yanulevskaya2008emotional} introduced an emotion categorization system predicated on evaluating local image statistics learned for each emotional category using a support vector machine. Alameda-Pineda \etal \cite{alameda2016recognizing} tackled recognizing emotions evoked by abstract paintings by employing a multi-label classifier. Lu \etal \cite{lu2012shape} delved into investigating shape features within images that impact the emotions elicited in humans. Zhao \etal \cite{zhao2014affective} explored the performance of various features across different image types within a multi-graph learning framework, subsequently fusing them for visual emotion recognition.

In contrast to hand-crafted methods, the deep representation learning approach has made significant improvements in visual emotion analysis. Chen \etal \cite{chen2014deepsentibank} introduced a visual sentiment concept classification network tailored to address biased training data comprising images with strong sentiment. You \etal \cite{you2015robust} devised a deep representation learning network equipped with innovative training strategies to mitigate the inherent noise in large-scale training datasets for visual emotion analysis. Rao \etal \cite{rao2019multi} developed a feature pyramid network to extract multi-level deep representations from visual emotion images. Addressing the fine-grained visual emotion regression task, Zhao \etal \cite{zhao2019pdanet} proposed a deep network integrating visual attention mechanisms into convolutional networks. Wei \etal \cite{wei2020learning} introduced a method for acquiring robust visual features for emotion analysis. Panda \cite{panda2018contemplating} conducted a comprehensive analysis of the existing visual emotion analysis benchmarks and explored the feasibility of training models directly using web data devoid of annotations.

 While deep representation learning-based visual emotion analysis outperforms hand-crafted methods significantly, these approaches often fail to harness the vital components inherent in most images, namely attributes. Diverging from prior deep representation approaches, Yang \etal \cite{yang2021stimuli} drew inspiration from the Stimuli-Organisms-Response model of psychological response \cite{Mehrabian2017CommunicationWords} in perceived emotion. They devised a stimuli-aware visual emotion analysis network capable of selecting stimuli and extracting distinct emotion features from various stimuli. Xu \etal \cite{xu2022mdan} dissected the affective gap into smaller gaps to address fine-grained emotion classification. Yang \etal \cite{yang2023emoset} introduced attribute-aware visual emotion recognition by leveraging low, mid, and high-level features to focus on diverse visual details from an input image.

 Drawing inspiration from Yang \etal \cite{yang2023emoset}, this paper embarks on an investigation into the realm of visual emotion representation learning, with a particular focus on the integration of four fundamental attributes: brightness, colorfulness, scene recognition, and facial expression recognition. The inclusion of these attributes stems from their pivotal roles in shaping the emotional perception of visual stimuli.

Firstly, incorporating the brightness attribute is grounded in its well-established significance within perceptual processing. Studies have consistently demonstrated the crucial influence of overall lighting levels in images on human emotional responses \cite{kurt2017modulation}. By considering brightness as a key attribute, we aim to illuminate its nuanced impact on visual emotion representation. The colorfulness of an image emerges as another critical attribute deserving attention. Research has indicated that the color composition of an image holds significant correlations with the elicited emotional responses \cite{ritchie2013perceived}. By delving into the complex relationship between colorfulness and emotional perception, we want insights into visual emotion representation.

Furthermore, scene recognition emerges as a compelling attribute to explore within the context of visual emotion analysis. Borosch \etal \cite{brosch2010perception} emphasized the importance of understanding the scene depicted in an image as a potent emotional stimulus. By integrating scene recognition into our study, we aim to unravel the emotional nuances embedded within diverse visual contexts. Lastly, facial expression recognition is an undeniable cornerstone of visual emotion analysis. Extensive research, notably by Ekman \cite{ekman1993facial}, underscores the profound impact of facial expressions on shaping the emotional experience of individuals. Through meticulous examination of facial expressions, we endeavor to elucidate their intricate role in visual emotion representation. 

\subsection{Scene Recognition}
Scene classification is a fundamental task in computer vision, aiming to automatically categorize images or videos into predefined classes or categories based on their visual content. This task involves a thorough analysis of diverse visual cues, including color, texture, shape, and spatial arrangement, to discern the contextual environment portrayed within the scene. With the dawn of deep representation learning methods, there has been a significant enhancement in the accuracy and efficiency of scene classification systems. This progress has enabled robust recognition of scenes within real-world environments.

Given the importance of scene understanding and recognition in computer vision, numerous methodologies have been proposed to develop effective scene representations. Global convolution network-based approaches, for instance, directly predict scene category probabilities from the entire scene image. Zuo \etal \cite{zuo2016learning} introduced hierarchical LSTM architectures to comprehend the contextual relationships between images and scene categories. Meanwhile, Xie \etal \cite{xie2016interactive} devised a global convolutional feature extraction network that integrates high-level visual context with low-level neuron responses. Rezanejad \etal \cite{rezanejad2019scene} demonstrated superior performance when utilizing the entire image as input for convolutional networks to capture essential information.

Considering the balance between simplicity and performance, our approach adopts a global scene recognition strategy by integrating the scene branch. This decision is informed by the effectiveness demonstrated by such methods in capturing the overarching context and facilitating accurate scene classification.

\subsection{Facial Expression Recognition}

Recognizing facial expressions holds importance in visual emotion analysis due to the expressiveness and informativeness of the human face in conveying emotions. Facial expressions offer cues about emotional state, encompassing happiness, sadness, anger, fear, surprise, and more. Given its practical significance in diverse domains, automatic facial expression analysis has garnered considerable attention from researchers \cite{Li2020}.

In recent years, facial expression recognition has made substantial progress, akin to the progress observed in scene recognition, primarily driven by deep representation learning techniques. Kaya \etal \cite{kaya2017video} explored expression recognition in diverse real-world settings, highlighting the performance of VGG-Face, initially trained for face recognition, over ImageNet in facial expression recognition tasks. Ng \etal \cite{ng2015deep} introduced a transfer learning approach for facial expression recognition, employing a two-stage process to leverage pre-trained models effectively.

In the context of visual emotion analysis, Yang \etal \cite{yang2021stimuli} developed Expression-Net, leveraging facial expression detection within emotional contexts. Yang \etal \cite{yang2023emoset} further advanced this area with a visual emotion network capable of detecting facial expressions directly from images without preprocessing. Inspired by Yang \etal \cite{yang2023emoset}, we incorporate the facial expression branch into our approach to recognizing facial expressions without needing to preprocess the image.

\section{Methodology}
\begin{figure*}
    \centering
    \includegraphics[width=\textwidth]{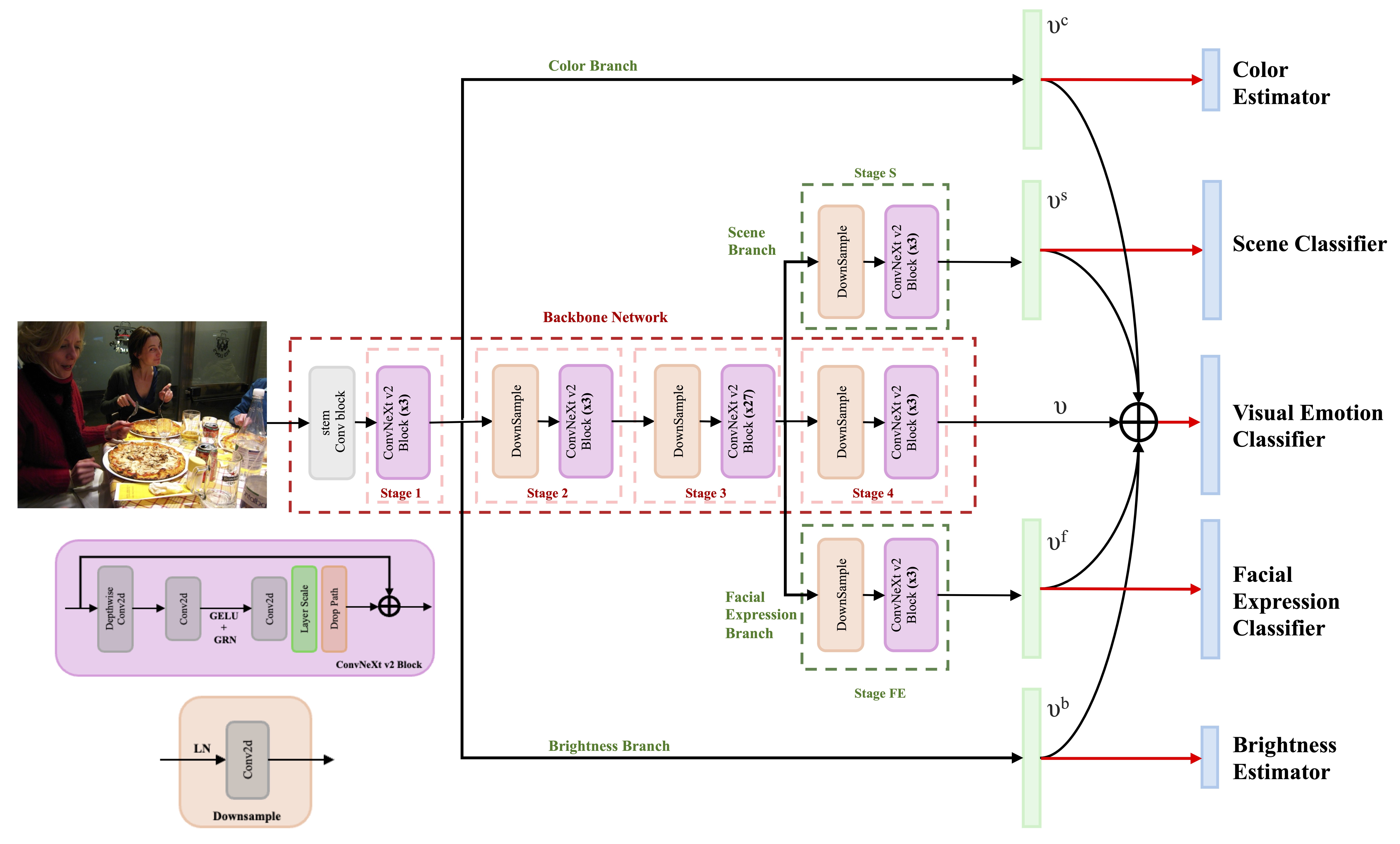}
    \caption{A4Net consists of one backbone network and four attribute branches. Specifically, the color branch is tasked to estimate the color intensity, and the brightness branch is employed to estimate brightness. The scene and facial expression branch is tasked to classify the image into the specific scene and facial expression classes. The feature vector from four branches is fused subsequently to classify visual emotion.}\label{fig:architecture}
\end{figure*}
\subsection{Overview}

This paper presents the attributes-aware visual emotion network called \textbf{A4Net}. Our approach, presented in Figure \ref{fig:architecture}, consists of four attribute branches, each proposed to extract specific visual cues from input images. These branches serve as specialized pathways to estimate key attributes such as color and brightness and classify details regarding scene understanding and facial expressions.

A4Net acts as a multi-label classification and estimator network \cite{han2023survey}, simultaneously tasked with multiple objectives. Specifically, it is designed to recognize visual emotions across a diverse spectrum of distinct classes. Moreover, it adeptly identifies various facial expressions, including six facial expression types, alongside an additional category dedicated to instances where no facial expression is detected. Furthermore, it categorizes scenes from 254 classes, with an added class for cases where the scene is unidentifiable. Additionally, A4Net can estimate the color and brightness of the input images, further enriching the understanding of visual emotion content.

The A4Net comprises of backbone \cite{woo2023convnext} and multiple branches, each developed to extract essential visual features. Collectively, these branches yield one-dimensional feature vectors of varying shapes, forming the backbone of visual emotion recognition.

\subsection{Color Branch}
        We extract features from the first stage of the backbone network for the color branch. Subsequently, these features undergo processing through a pre-estimator layer, with the composition as follows:
        \begin{equation}\label{eq:color}
        v^{c}=FC\lbrack ~NORM \lbrack ~GAP({CNB}^{1}_{-1}(v^{2}))\rbrack \rbrack,
        \end{equation}  
        here, ${CNB}^{1}_{-1}$ represents output of \textit{third} ConvNeXt-V2 \cite{woo2023convnext} block of stage-1 which is 1-dimensional feature vector of shape $128$.  $v^{2}$ represents the output feature vector from the \textit{second} ConvNeXt-V2 \cite{woo2023convnext} block of stage-1. Following the feature vector extraction, the vector undergoes global average pooling ($GAP$), then layer normalization ($NORM$), and finally passes through a fully connected layer (FC), resulting in a 1-dimensional vector of shape $1024$. $v^{c}$ represent  color feature vector. 
        
        The color feature vector $v^{c}$ is then passed through a linear layer with one output node for color regression, resulting in $\hat{y^{c}}$. 
        
        We employ mean square error loss for color estimation and write as follows:
        \begin{equation}\label{eq: colorbranch}
        \mathcal{L}_{\text{C}} = \frac{1}{n} \sum_{i=1}^{n} (y^{c}_i - \hat{y^{c}_i})^2,
        \end{equation}
        where, $n$ is number of images, $y^{c}_i$ represents the ground truth  color value, $\hat{y^{c}_i}$
        denotes the predicted  color value  at $i^{th}$ images.

\subsection{Brightness Branch }
        Similar to the Color branch, For the brightness branch, we extract the features from stage 1 of the backbone network. Following the extraction, the features are passed through a pre-estimator layer and are composed as follows: 
        \begin{equation}\label{eq:brightness}
        v^{b}=FC\lbrack ~NORM \lbrack ~GAP({CNB}^{1}_{-1}(v^{2}))\rbrack \rbrack,
        \end{equation}  
        here, ${CNB}^{1}_{-1}$ represents output of \textit{third} ConvNeXt-V2 \cite{woo2023convnext}  block of stage-1 which is 1-dimensional feature vector of shape $128$. $v^{2}$ represents the output feature vector from the \textit{second} ConvNextv-2 \cite{woo2023convnext} block of stage-1. Following the feature vector extraction, the vector undergoes global average pooling ($GAP$), then layer normalization ($NORM$), and finally passes through a fully connected layer (FC), resulting in a 1-dimensional vector of shape $1024$. $v^{b}$ represent brightness feature vector. 
        
        The brightness feature vector $v^{b}$ is then passed through a linear layer with one output node for brightness regression, resulting in $\hat{y^{b}}$. 
        Similar to the color branch, we employ mean square error loss for brightness estimation and write as follows:
        \begin{equation}\label{eq:brightnessloss}
        \mathcal{L}_{\text{B}} = \frac{1}{n} \sum_{i=1}^{n} (y^{b}_i - \hat{y^{b}_i})^2,
        \end{equation}
        where, $n$ is number of images, $y^{b}_i$ represents the ground truth brightness value, $\hat{y^{b}_i}$
        denotes the predicted brightness value  at $i^{th}$ images.

\subsection{Scene Branch}
        Due to the complex nature of scene representation, we opt to extract features from the final ConvNeXt-V2 \cite{woo2023convnext} block of stage-3. Like the preceding ConvNeXt-V2 \cite{woo2023convnext} blocks, the outputs are 1-dimensional feature vectors, albeit with a shape of 512. These extracted feature vectors are fed into a dedicated block termed \textbf{Stage S} for scene classification. Stage S comprises a DownSample module followed by three ConvNeXt-V2 \cite{woo2023convnext} blocks, mirroring the configuration of backbone Networks' Stage-4. Furthermore, Stage S is initialized with the weights inherited from the backbone Network Stage 4.
        
        Following the feature extraction, the features are passed through multiple layers, which are composed as follows: 
        \begin{equation}\label{eq:scene}
        v^{s}=FC\lbrack ~NORM \lbrack ~GAP({CNB}^{S}_{-1}(v^{27}))\rbrack \rbrack,
        \end{equation}  
        here, ${CNB}^{S}_{-1}$ represents output of third ConvNeXt-V2 \cite{woo2023convnext}  block of stage-S which is 1-dimensional feature vector of shape $512$. $v^{27}$ represents the feature vector generated by the final layer of ConvNeXt-V2 \cite{woo2023convnext}  block of stage-3. Following the feature vector extraction, the vector undergoes global average pooling ($GAP$), then layer normalization ($NORM$), and finally passes through a fully connected layer (FC), resulting in a 1-dimensional vector of shape $1024$. $v^{s}$ represent scene feature vector. 
        
        The scene feature vector $v^{s}$ is then passed through a linear layer with 255 (254+1) output nodes for scene classification, resulting in $\hat{y^{s}}$. 
        
        For scene classification, we employ cross-entropy loss and write as:
        \begin{equation}\label{scenebranch}
        \mathcal{L}_{\text{S}} = - \frac{1}{n} \sum_{i=1}^{n} \left( y^{s}_i \log(\hat{y^{s}}_i) + (1 - y^{s}_i) \log(1 - \hat{y^{s}}_i) \right),
        \end{equation}
        where, $n$ is number of images, $y^{s}_i$ represents the ground truth scene class, $\hat{y^{c}_i}$
        denotes the predicted scene class  at $i^{th}$ images. We have also added a class for \textit{unknown scenes}. 

\subsection{Facial Expression Branch}
    Similarly, following the scene branch,  we extract the features from the \textit{last} ConvNeXt-V2 \cite{woo2023convnext} block of stage-3 for the facial expression branch. The outputs are 1-dimensional feature vectors of shape 512. Like the scene classifier, the extracted feature vector is passed through a separate block called \textbf{Stage FE}, similar to Stage S. 
    
    Following the feature extraction, for facial expression, features are passed through multiple layers, which are composed as follows: 
    \begin{equation}\label{eq:facialexpression}
    v^{f}=FC\lbrack ~NORM \lbrack ~GAP({CNB}^{FE}_{-1}(v^{27}))\rbrack \rbrack
    \end{equation}  
    here, ${CNB}^{FE}_{-1}$ represents output of third ConvNeXt-V2 \cite{woo2023convnext} block of stage-FE which is 1-dimensional feature vector of shape $512$. $v^{27}$ represents the feature vector generated by the final layer of ConvNeXt-V2 \cite{woo2023convnext} block of stage-3. Following the feature vector extraction, the vector undergoes global average pooling ($GAP$), then layer normalization ($NORM$), and finally passes via a fully connected layer (FC), resulting in a vector of 1-dimensional of shape $1024$. $v^{f}$ represents the facial expression feature vector. 
    
    The scene feature vector $v^{f}$ is then passed through a linear layer with 7 (6+1) output nodes for facial expression classification, resulting in $\hat{y^{fe}}$. 
    
    For facial expression classification, we employ cross-entropy loss and write as:
    \begin{equation}\label{eq:facialexpressionbranch}
        \mathcal{L}_{\text{FE}} = - \frac{1}{n} \sum_{i=1}^{n} \left( y^{fe}_i \log(\hat{y^{fe}}_i) + (1 - y^{fe}_i) \log(1 - \hat{y^{fe}}_i) \right)
    \end{equation}
    where, $n$ is number of images, $y^{fe}_i$ represents the ground truth facial expression class, $\hat{y^{fe}_i}$
    denotes the predicted facial expression class  at $i^{th}$ images. We have added a class for \textit{unknown facial expressions or no face} in the image. 

\subsection{Visual Emotion Classifier}
    At the core of A4Net lies the backbone network \cite{woo2023convnext}, which serves as a cornerstone for all its branches, as depicted in Figure \ref{fig:architecture}. The feature vectors obtained from the color estimator, brightness estimator, scene classifier, and facial expression classifier branches heavily rely on the knowledge learned by the backbone network. These branches combine their respective feature vectors with the output $v$ from the final block of Stage 4 to make predictions about the emotion class, as outlined in equation \ref{eq:all}.
    
    \begin{equation}\label{eq:all}
    \hat{y}=FC\lbrack v + w^{c}.v^{c} + w^{b}.v^{b} + w^{s}.v^{s}+ w^{f}.v^{f}\rbrack,
    \end{equation}  
    here, $w^{c}$, $w^{b}$, $w^{s}$, and $w^{f}$ are trainable parameters for controlling the weight of different branches. 
    
    For the visual emotion classifier, we employ cross-entropy loss and write as:
    \begin{equation}\label{eq:visualemotionclassifier}
    \mathcal{L}_{\text{VE}} = - \frac{1}{n} \sum_{i=1}^{n} \left( y_i \log(\hat{y}_i) + (1 - y_i) \log(1 - \hat{y}_i) \right)
    \end{equation}

    Based on the multi-label classification and regression task for A4Net,  an overall objective is written as follows:
    \begin{equation}\label{eq:objective}
    \displaystyle \mathop{\mathrm{argmin}} \big (\mathcal{L}_{VE} + w_{B}\mathcal{L}_{B} + w_{C}\mathcal{L}_{C} + w_{S}\cdot\mathcal{L}_{S} +w_{FE} \cdot\mathcal{L}_{FE}\big)
    \end{equation}  
    
    where $w_{B}$, $w_{C}$, $w_{S}$, and $w_{FE}$ are trainable parameters for regularization for focusing on the important visual features. 

\section{Experiments}

    \subsection{Datasets}
        We evaluate the performance of A4Net on four different visual emotion datasets. 
        \begin{itemize}
        
             \item \textbf{EmoSet}: Emoset \cite{yang2023emoset} stands as a comprehensive visual emotion dataset, boasting a vast collection of 3.3 million images, each endowed with rich attributes. These attributes encompass brightness, colorfulness, scene context, human actions, facial expressions, and object characteristics. In configuring our experiments for visual emotion recognition, we align with the methodology outlined by Yang et al. \cite{yang2023emoset}, allocating proportions of $80\%$, $5\%$, and $15\%$ for training, validation, and test sets, respectively. 
           \item \textbf{EMOTIC}: The EMOTIC dataset \cite{kosti2019context} is a compilation of images sourced from various sources that include MSCOCO \cite{lin2014microsoft}, Ade20K \cite{zhou2017scene}, and additional images obtained through Google search. This dataset features images capturing individuals in natural settings,   annotated to depict their emotions. In total, the dataset comprises 18,000 images. In this study, we specifically focus on evaluating the performance of our model solely on the EMOTIC-I(mage) subset. We adhere to the training and evaluation protocols delineated in \cite{kosti2019context} to ensure consistency.
            
            \item \textbf{SE30K8}: The SE30K8 dataset \cite{wei2020learning} comprises a collection of 33,000 images, each annotated using Amazon Mechanical Turk (AMT). We adopt the training, validation, and testing procedures outlined in our experimental setup in \cite{wei2020learning}.

            \item \textbf{UnBiasEmo}: The UnBiasEmo dataset \cite{panda2018contemplating} encompasses 3,000 images sourced from Google, capturing various emotions associated with identical entities to mitigate object bias. Each image is labeled with six emotional classes. We adhere to the training and testing methodologies outlined by Panda \etal \cite{panda2018contemplating} to maintain consistency.
        \end{itemize}

    \subsection{Baselines}
        To showcase the effectiveness of A4Net, we conduct a comparative analysis with several baseline models using the EmoSet Dataset. The baselines include traditional convolutional networks and visual emotion analysis networks. Additionally, in line with the findings of Yang \etal \cite{yang2023emoset}, we evaluate our performance against the attribute-aware convolutional network. Specifically, for traditional convolutional networks, we compare against \textit{AlexNet} \cite{Krizhevsky2017ImageNetNetworks}, \textit{VGGNet-16} \cite{Simonyan2015VeryRecognition}, \textit{ResNet-50} \cite{he2016deep}, and \textit{DenseNet-121} \cite{huang2017densely}. Furthermore, we examine the attribute-aware visual emotion analysis models proposed by Yang \etal \cite{yang2023emoset} that contain three branches to extract visual information at low, medium, and high levels. Table \ref{emosetaccuracy} presents the performance results of four different attribute-module attached models, namely \textit{AlexNet with three levels} \cite{yang2023emoset}, \textit{VGGNet-16 with three levels} \cite{yang2023emoset}, \textit{ResNet-50 with three levels} \cite{yang2023emoset}, and \textit{DenseNet-121 with three levels} \cite{yang2023emoset}.

        Furthermore, we thoroughly compare the A4Net model trained on the EmoSet dataset with multiple visual emotion networks. Among these methods, \textit{WSCNet} \cite{yang2018weakly} introduces a weakly supervised coupled network adept at selecting relevant soft proposals based on weak annotations, such as global image labels. Meanwhile, \textit{StyleNet} \cite{zhang2019exploring}l earns content representations from higher layers of the network and combines style information from different layers, thus achieving a holistic understanding of visual content.
        
        On the other hand, \textit{PDANet} \cite{zhao2019pdanet} presents an approach by integrating attention mechanisms directly into the convolutional network while adhering to emotional polarity constraints to ensure consistent emotional representations. Additionally, \textit{Stimuli-aware} \cite{yang2021stimuli} mimics the human evocation process through a multi-stage approach, providing a deeper insight into the emotional response elicited by visual stimuli. Lastly, \textit{MDAN} \cite{xu2022mdan} utilizes both bottom-up and top-down branches to capture global and level-wise discriminative features using multiple classifiers.
    \subsection{Implementation Details}
        For EmoSet,  A4Net's backbone network uses ConvNeXt-V2 \cite{woo2023convnext}, pre-trained on ImageNet \cite{deng2009imagenet}. Both scene and facial expression branches are identical to ImageNet \cite{deng2009imagenet}  pre-trained stage-4. All branches' penultimate fully connected layer ($FC$) is set to a $1024$ dimensional feature vector. We perform random image crop to 224x224 and horizontal flips randomly. We use a weight decay of 0.0001. A batch size of 80 and a learning rate of 0.000003 is used. A4Net is trained for 20 Epochs for EmoSet.

    \subsection{Evaluation of learned visual features}
        We assess the efficacy of the features trained on EmoSet by employing them for visual emotion recognition tasks on EMOTIC, SE30K8, and UnBiasEmo datasets. Following a methodology proposed by Wei \etal \cite{wei2020learning}, we utilize A4Net trained on the EmoSet dataset for image feature extraction. These features are directly applied without fine-tuning the target task. We employ a straightforward linear classifier for emotion categorization to gauge the effectiveness of the visual features extracted by A4Net.
        
        We maintain all layers of A4Net in a frozen state and substitute the last fully connected layer of the Visual Emotion Classifier within A4Net with a new trainable layer designed to map the learned features to the output classes of the target dataset. This newly added layer is trained exclusively on the target dataset. For the EMOTIC dataset comprising 26 classes, we opt for a batch size of 80, set the learning rate to 0.002, and train the model for 30 epochs. Given that the EMOTIC dataset encompasses multiple labels for each image, we utilize binary cross-entropy loss during training.
        
        Similarly, for the SE30K8 dataset featuring eight classes, we employ a batch size of 80, a learning rate of 0.003, and conduct training for 30 epochs. The loss function employed for SE30K8 is identical to that used for the EmoSet dataset.
        
        Lastly, for the UnBiasedEmo dataset, which comprises six emotion classes, we adopt a batch size of 2 and set the learning rate to 0.00007. Consistent with the SE30K8 and EmoSet datasets, we employ cross-entropy loss during training.
        
        \begin{figure*}[ht!]
            \centering
            \includegraphics[width=\textwidth]{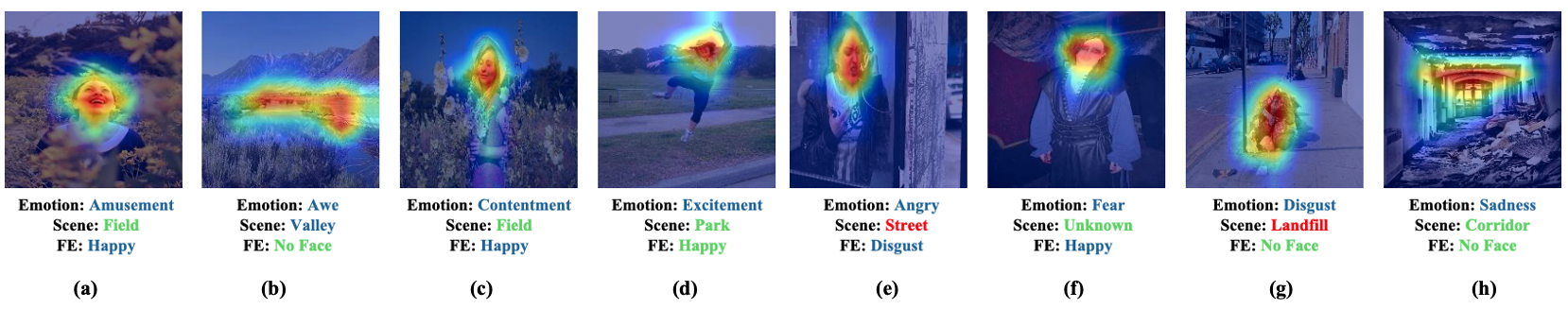}
            \caption{Visualization using GradCAM of A4Net trained on the EmoNet Dataset. Words highlighted in \textit{blue} indicate correct classification. Words highlighted in \textit{red} indicate cases where A4Net recognizes the wrong class. Words highlighted in \textit{green} represent classes not present in the test dataset. (Best viewed in Color)}\label{fig:emosetvis}
        \end{figure*}
        \begin{figure*}[t]
            \centering
            \includegraphics[width=\textwidth]{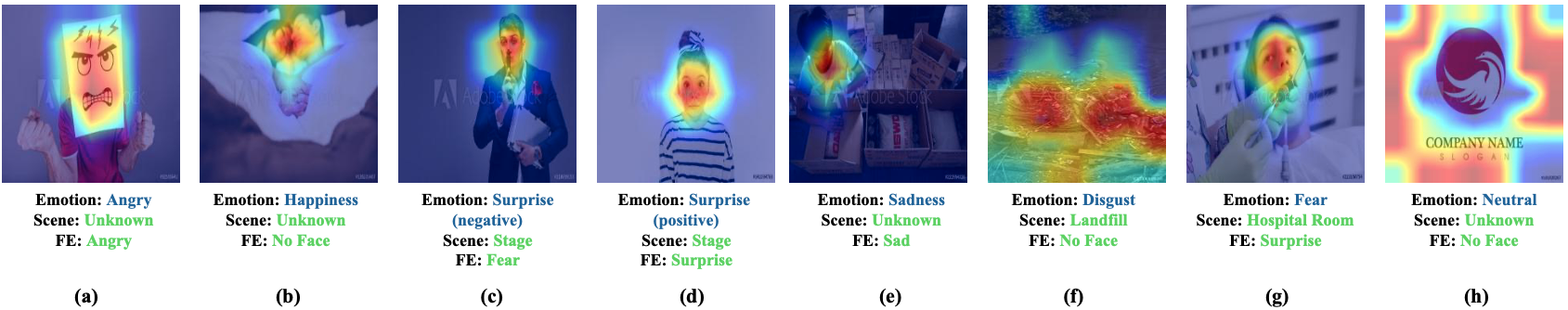}
            \caption{GradCAM visualization showcasing performance of A4Net  on the SE30K8 Dataset. Words highlighted in \textit{blue} denote correct classifications. Instances where A4Net identifies classes not present in the test dataset are highlighted in \textit{green}. (Best viewed in Color)}\label{fig:se30k8vis}
        \end{figure*}
    

\subsection{Comparisons}
Performance evaluations comparing the proposed A4Net with state-of-the-art approaches are based on accuracy metrics for the EmoSet, UnBiasEmo, and SE30K8 datasets and mean Average Precision (mAP) for the EMOTIC-I dataset. The results are presented in Tables \ref{emosetaccuracy} and \ref{otheraccuracy}. Analyzing these findings allows us to draw the following conclusions:

\begin{table}
    \caption{Top-1 accuracy comparison of various visual emotion recognition on EmoSet Dataset.}\label{emosetaccuracy}
    \centering
    \begin{tabularx}{\columnwidth}{X|X}
        \hline
        Models                 & Top-1 Accuracy $(\%)$    \\
        \hline
        AlexNet \cite{Krizhevsky2017ImageNetNetworks} & 67.8 \\
        VGGNet-16 \cite{Simonyan2015VeryRecognition}     & 72.27       \\
        ResNet-50 \cite{he2016deep}        & 74.04 \\
        DensNet-121 \cite{huang2017densely}        & 72.32 \\
        WSCNet \cite{yang2018weakly}        & 76.32 \\
        StyleNet \cite{zhang2019exploring}        & 77.11 \\
        PDANet \cite{zhao2019pdanet}        & 76.95 \\
        Stimuli-aware \cite{yang2021stimuli}        & 78.4 \\
        MDAN \cite{xu2022mdan}        & 75.75 \\
        AlexNet with three levels \cite{yang2023emoset} & 70.09\\
        VGGNet-16 with three levels\cite{yang2023emoset}     & 74.76      \\
        ResNet-50 with three levels\cite{yang2023emoset}        & 76.60 \\
        DensNet-121 with three levels\cite{yang2023emoset}        & 74.94 \\
        \textbf{A4Net (ours)} & \textbf{85.0}\\
        \hline
    \end{tabularx}
\end{table}

\begin{table}
    \caption{Mean Average Precision (mAP) and Top-1 accuracy comparison of baseline and previous State-of-the-art (SOTA) method with A4Net on EMOTIC-I, UnBiasEmo and SE30K8 Dataset.}\label{otheraccuracy}
    \centering
     \begin{tabularx}{\columnwidth}{X|X|X|X}
        \hline
        Models                 & EMOTIC-I (mAP) &UnBiasEmo &SE30K8   \\
        \hline
        ResNet-50 \cite{he2016deep}      &26.03  & 60.26 & 52.52\\
        SOTA \cite{wei2020learning}    &30.96  & 81.45 & \textbf{69.78}\\
        \textbf{A4Net (ours)} & \textbf{32.77} & \textbf{82.4}&64.69\\
        \hline
    \end{tabularx}
\end{table}
\begin{figure}[ht!]
    \centering
    \includegraphics[width=\columnwidth]{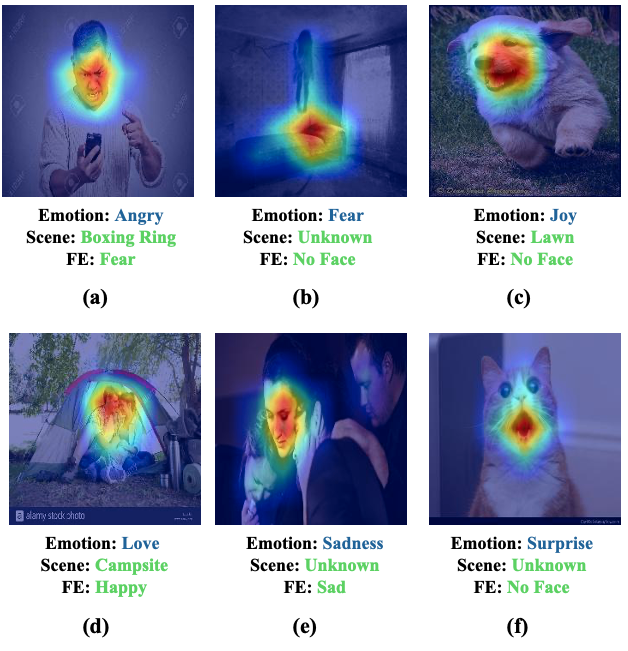}
    \caption{GradCAM visualization of A4Net trained on UnBiasEmo Dataset. The word in \textit{blue} represents that A4Net can be in the true class correctly. The word in \textit{green} indicates the class not in the test dataset. (Best viewed in Color)}\label{fig:unbiasemovis}
\end{figure}

\begin{enumerate}
    \item Traditional convolutional networks, which conduct feature extraction and subsequently feed these features into a standard classifier, exhibit inferior performance. This decline in performance can be attributed to an affective gap, wherein directly utilizing extracted features may prove inconsistent with the visual emotions being analyzed, as they may encompass abstract concepts.
     \item In the majority of cases, employing A4Net trained on EmoSet for transfer learning on other visual emotion datasets yields commendable performance. The outcomes depicted in Table \ref{otheraccuracy} underscore the ability of A4Net to acquire generalized visual emotion features.
     \item The proposed A4Net demonstrates superior performance. The EmoSet dataset shows a notable $5.1\%$ difference in top-1 accuracy between our proposed A4Net and the previous state-of-the-art model \cite{yang2021stimuli}. Similarly, on the EMOTIC-I and UnBiasEmo datasets, A4Net outperforms the previous state-of-the-art model \cite{wei2020learning}. However, on the SE30K8 dataset, the performance of A4Net  is comparatively impacted compared to the state-of-the-art model \cite{wei2020learning}. It is essential to recognize that the state-of-the-art model \cite{wei2020learning} is pre-trained on the StockEmotion dataset \cite{wei2020learning}, consisting of 1.17 million images with 690 keywords as classes. Subsequently, the model is fine-tuned on the SE30K8 dataset, a subset of StockEmotion, with manually annotated labels. Due to this pre-training and fine-tuning process on the same input image distribution, the state-of-the-art model \cite{wei2020learning} tends to outperform A4Net.
     \item Table \ref{abla} presents the effect of different attributes in the overall visual emotion analysis. It shows that with all four attributes, A4Net performs better and improves the attribute branches.    
\end{enumerate}

\section{Visualization}
\begin{table}
    \caption{Ablation Study effect on A4Net with different attributes: B (Brightness), C (Color), S (Scene), and F(Facial Expression). }\label{abla}
    \centering
     \begin{tabularx}{\columnwidth}{X|X|X|X|X|X}
        \hline
              &Emotion ($\%$) &B(MSE)&C(MSE)&S&F \\
        \hline
        B           &82.03        &0.022  &   -  &    -  &    - \\
        C           &82.01        &- &0.041&-&- \\
        S           &81.08        &-&-&\textbf{65.08}&- \\
        F           &82.88        &-&-&-&81.75 \\
        S+F         &82.91        &-&-&62.79&\textbf{85.10} \\
        B+S+F       &83.91    &0.018&-&64.74&82.01\\
        C+S+F       &83.83    &-&0.033&64.92&82.60 \\
        B+C+S+F~     &\textbf{85.05}  &\textbf{0.009}&\textbf{0.001}&65.02&82.92 \\
        \hline
        
        \hline
    \end{tabularx}
\end{table}

To demonstrate the interpretability of A4Net, we utilize the heatmaps generated by GradCAM \cite{selvaraju2017grad} to visualize the learned activations. As depicted in Figures \ref{fig:emosetvis}, \ref{fig:se30k8vis}, and \ref{fig:unbiasemovis}, we observe that activation maps of A4Net effectively pinpoint regions of the image relevant to visual emotions. For instance, in Figure \ref{fig:emosetvis}(a), A4Net accurately classifies the image as \textit{Amusement} by focusing on the person displaying a \textit{Happy} facial expression. Similarly, in Figure \ref{fig:emosetvis}(h), A4Net correctly identifies the image depicting sadness, with its focus directed towards the corridor, recognizing the absence of faces or facial expressions.

Likewise, Figure \ref{fig:se30k8vis} visualizes the activation map generated by A4Net trained on the SE30K8 Dataset. Notably, the SE30K8 dataset does not include attribute annotations; however, the ability of A4Net  to recognize various attributes still significantly correlates with the depicted emotions in the images. For example, in Figure \ref{fig:se30k8vis}(a), A4Net accurately identifies the presence of \textit{angry} emotion by focusing on facial expressions of \textit{anger}, despite the absence of explicit attribute annotations in the dataset. This underscores the capability of A4Net  to discern relevant visual cues for emotion recognition, even in datasets lacking specific attribute labels.

In Figure \ref{fig:unbiasemovis}(d), A4Net identifies the emotional content of the image as \textit{Love}. It focuses on two individuals depicted within a \textit{Campsite} scene, both displaying \textit{Happy} facial expressions, effectively capturing the essence of \textit{Love} portrayed in the image. Furthermore, in Figures \ref{fig:unbiasemovis}(c) and (f), the emotions conveyed are \textit{Joy} and \textit{Surprisee}, respectively. A4Net discerns the absence of any human facial expressions in these images, indicating its ability to recognize emotions through other visual cues beyond facial expressions. This demonstrates the robustness of A4Net in accurately interpreting emotions across diverse visual contexts.

Figure \ref{fig:emosetvis},\ref{fig:se30k8vis}, and \ref{fig:unbiasemovis} visually verifies that the mapping of the scene and facial expression attributes are crucial in better visual emotion analysis.

\section{Conclusion and Future Directions}
This paper addresses the challenges of visual emotion analysis by leveraging various attributes, such as the colorfulness or brightness of an image, the understanding of the scene in an image, or the presence of facial expressions on a given image. 
This paper addresses the challenge by developing a deep representation learning network named \textbf{A4Net}. A4Net integrates four key attributes - brightness (\textit{Attribute 1}), colorfulness (\textit{Attribute 2}), scene (\textit{Attribute 3}), and facial expression (\textit{Attribute 4}) - by combining and jointly training all attribute recognition and visual emotion recognition components. Extensive experiments and visualizations conducted on the EmoSet, EMOTIC, SE30K8, and UnBiasEmo datasets illustrate that A4Net surpasses existing approaches for visual emotion recognition.

A4Net fuses only four different attributes. However, other attributes could be considered important components of visual emotion analysis, such as objects or human activities in a given image. In future work, we will address the permutational relationship between various attributes that evoke emotion in images. For instance, we are interested in understanding the interplay between \textit{scene + facial expression}, \textit{scene + human activity}, \textit{object + facial expression}, \etc ~that can evoke emotion in images. Furthermore, most of the images used in training and testing are natural \ie, mostly taken from daily human surroundings; how the deep representation learning model with different attributes can understand visual emotion from diverse abstract images such as Figure \ref{fig:se30k8vis}(h) would be an interesting investigation.


\bibliographystyle{ieeetr} 
\bibliography{references_edit} 

\end{document}